\documentclass[10pt,twocolumn,letterpaper]{article}

\usepackage[pagenumbers]{wacv} 

\usepackage{graphicx}
\usepackage{amsmath}
\usepackage{amssymb}
\usepackage{booktabs}
\usepackage{multirow, makecell}
\usepackage{caption}
\usepackage{textcomp}
\usepackage{xcolor, colortbl}
\usepackage{mathtools}
\usepackage{appendix}
\usepackage{enumitem}
\usepackage{eucal}
\usepackage{amsfonts}
\usepackage{xspace}

\usepackage[pagebackref,breaklinks,colorlinks]{hyperref}

\usepackage[capitalize]{cleveref}
\crefname{section}{Sec.}{Secs.}
\Crefname{section}{Section}{Sections}
\Crefname{table}{Table}{Tables}
\crefname{table}{Tab.}{Tabs.}

\def\wacvPaperID{887} 
\def\confName{WACV}
\def\confYear{2024}

\def\Approach{LP-OVOD}

\usepackage{mathtools}

\begin{document}
\def\mA{\mathcal{A}}
\def\mB{\mathcal{B}}
\def\mC{\mathcal{C}}
\def\mD{\mathcal{D}}
\def\mE{\mathcal{E}}
\def\mF{\mathcal{F}}
\def\mG{\mathcal{G}}
\def\mH{\mathcal{H}}
\def\mI{\mathcal{I}}
\def\mJ{\mathcal{J}}
\def\mK{\mathcal{K}}
\def\mL{\mathcal{L}}
\def\mM{\mathcal{M}}
\def\mN{\mathcal{N}}
\def\mO{\mathcal{O}}
\def\mP{\mathcal{P}}
\def\mQ{\mathcal{Q}}
\def\mR{\mathcal{R}}
\def\mS{\mathcal{S}}
\def\mT{\mathcal{T}}
\def\mU{\mathcal{U}}
\def\mV{\mathcal{V}}
\def\mW{\mathcal{W}}
\def\mX{\mathcal{X}}
\def\mY{\mathcal{Y}}
\def\mZ{\mathcal{Z}} 

\def\bbN{\mathbb{N}} 
\def\bbR{\mathbb{R}} 
\def\bbP{\mathbb{P}} 
\def\bbQ{\mathbb{Q}} 
\def\bbE{\mathbb{E}}

\def\1n{\mathbf{1}_n}
\def\0{\mathbf{0}}
\def\1{\mathbf{1}}

\def\A{{\bf A}}
\def\B{{\bf B}}
\def\C{{\bf C}}
\def\D{{\bf D}}
\def\E{{\bf E}}
\def\F{{\bf F}}
\def\G{{\bf G}}
\def\H{{\bf H}}
\def\I{{\bf I}}
\def\J{{\bf J}}
\def\K{{\bf K}}
\def\L{{\bf L}}
\def\M{{\bf M}}
\def\N{{\bf N}}
\def\O{{\bf O}}
\def\P{{\bf P}}
\def\Q{{\bf Q}}
\def\R{{\bf R}}
\def\S{{\bf S}}
\def\T{{\bf T}}
\def\U{{\bf U}}
\def\V{{\bf V}}
\def\W{{\bf W}}
\def\X{{\bf X}}
\def\Y{{\bf Y}}
\def\Z{{\bf Z}}

\def\a{{\bf a}}
\def\b{{\bf b}}
\def\c{{\bf c}}
\def\d{{\bf d}}
\def\e{{\bf e}}
\def\f{{\bf f}}
\def\g{{\bf g}}
\def\h{{\bf h}}
\def\i{{\bf i}}
\def\j{{\bf j}}
\def\k{{\bf k}}
\def\l{{\bf l}}
\def\m{{\bf m}}
\def\n{{\bf n}}
\def\o{{\bf o}}
\def\p{{\bf p}}
\def\q{{\bf q}}
\def\r{{\bf r}}
\def\s{{\bf s}}
\def\t{{\bf t}}
\def\u{{\bf u}}
\def\v{{\bf v}}
\def\w{{\bf w}}
\def\x{{\bf x}}
\def\y{{\bf y}}
\def\z{{\bf z}}

\def\balpha{\mbox{\boldmath{$\alpha$}}}
\def\bbeta{\mbox{\boldmath{$\beta$}}}
\def\bdelta{\mbox{\boldmath{$\delta$}}}
\def\bgamma{\mbox{\boldmath{$\gamma$}}}
\def\blambda{\mbox{\boldmath{$\lambda$}}}
\def\bsigma{\mbox{\boldmath{$\sigma$}}}
\def\btheta{\mbox{\boldmath{$\theta$}}}
\def\bomega{\mbox{\boldmath{$\omega$}}}
\def\bxi{\mbox{\boldmath{$\xi$}}}
\def\bnu{\mbox{\boldmath{$\nu$}}}                                  
\def\bphi{\mbox{\boldmath{$\phi$}}}
\def\bmu{\mbox{\boldmath{$\mu$}}}

\def\bDelta{\mbox{\boldmath{$\Delta$}}}
\def\bOmega{\mbox{\boldmath{$\Omega$}}}
\def\bPhi{\mbox{\boldmath{$\Phi$}}}
\def\bLambda{\mbox{\boldmath{$\Lambda$}}}
\def\bSigma{\mbox{\boldmath{$\Sigma$}}}
\def\bGamma{\mbox{\boldmath{$\Gamma$}}}
                                  
\newcommand{\myprob}[1]{\mathop{\mathbb{P}}_{#1}}

\newcommand{\myexp}[1]{\mathop{\mathbb{E}}_{#1}}

\newcommand{\mydelta}[1]{1_{#1}}

\newcommand{\myminimum}[1]{\mathop{\textrm{minimum}}_{#1}}
\newcommand{\mymaximum}[1]{\mathop{\textrm{maximum}}_{#1}}    
\newcommand{\mymin}[1]{\mathop{\textrm{minimize}}_{#1}}
\newcommand{\mymax}[1]{\mathop{\textrm{maximize}}_{#1}}
\newcommand{\mymins}[1]{\mathop{\textrm{min.}}_{#1}}
\newcommand{\mymaxs}[1]{\mathop{\textrm{max.}}_{#1}}  
\newcommand{\myargmin}[1]{\mathop{\textrm{argmin}}_{#1}} 
\newcommand{\myargmax}[1]{\mathop{\textrm{argmax}}_{#1}} 
\newcommand{\myst}{\textrm{s.t. }}

\newcommand{\denselist}{\itemsep -1pt}
\newcommand{\sparselist}{\itemsep 1pt}

\definecolor{pink}{rgb}{0.9,0.5,0.5}
\definecolor{purple}{rgb}{0.5, 0.4, 0.8}   
\definecolor{gray}{rgb}{0.3, 0.3, 0.3}
\definecolor{mygreen}{rgb}{0.2, 0.6, 0.2}

\newcommand{\cyan}[1]{\textcolor{cyan}{#1}}
\newcommand{\red}[1]{\textcolor{red}{#1}}  
\newcommand{\blue}[1]{\textcolor{blue}{#1}}
\newcommand{\magenta}[1]{\textcolor{magenta}{#1}}
\newcommand{\pink}[1]{\textcolor{pink}{#1}}
\newcommand{\green}[1]{\textcolor{green}{#1}} 
\newcommand{\gray}[1]{\textcolor{gray}{#1}}    
\newcommand{\mygreen}[1]{\textcolor{mygreen}{#1}}    
\newcommand{\purple}[1]{\textcolor{purple}{#1}}       

\definecolor{greena}{rgb}{0.4, 0.5, 0.1}
\newcommand{\greena}[1]{\textcolor{greena}{#1}}

\definecolor{bluea}{rgb}{0, 0.4, 0.6}
\newcommand{\bluea}[1]{\textcolor{bluea}{#1}}
\definecolor{reda}{rgb}{0.6, 0.2, 0.1}
\newcommand{\reda}[1]{\textcolor{reda}{#1}}

\def\changemargin#1#2{\list{}{\rightmargin#2\leftmargin#1}\item[]}
\let\endchangemargin=\endlist
                                               
\newcommand{\cm}[1]{}

\newcommand{\mhoai}[1]{{\color{magenta}\textbf{[MH: #1]}}}

\newcommand{\mtodo}[1]{{\color{red}$\blacksquare$\textbf{[TODO: #1]}}}
\newcommand{\myheading}[1]{\vspace{1ex}\noindent \textbf{#1}}
\newcommand{\htimesw}[2]{\mbox{$#1$$\times$$#2$}}


\newif\ifshowsolution
\showsolutiontrue

\ifshowsolution  
\newcommand{\Comment}[1]{\paragraph{\bf $\bigstar $ COMMENT:} {\sf #1} \bigskip}
\newcommand{\Solution}[2]{\paragraph{\bf $\bigstar $ SOLUTION:} {\sf #2} }
\newcommand{\Mistake}[2]{\paragraph{\bf $\blacksquare$ COMMON MISTAKE #1:} {\sf #2} \bigskip}
\else
\newcommand{\Solution}[2]{\vspace{#1}}
\fi

\newcommand{\truefalse}{
\begin{enumerate}
	\item True
	\item False
\end{enumerate}
}

\newcommand{\yesno}{
\begin{enumerate}
	\item Yes
	\item No
\end{enumerate}
}

\newcommand{\Sref}[1]{Sec.~\ref{#1}}
\newcommand{\Eref}[1]{Eq.~(\ref{#1})}
\newcommand{\Fref}[1]{Fig.~\ref{#1}}
\newcommand{\Tref}[1]{Table~\ref{#1}}

\definecolor{mydarkblue}{rgb}{0,0.08,1}
\definecolor{mydarkgreen}{rgb}{0.02,0.6,0.02}
\definecolor{myred}{rgb}{1.0,0.0,0.0}
\definecolor{mydarkred}{rgb}{0.9,0.1,0.0}
\definecolor{babyblue}{rgb}{0.54, 0.81, 0.94}
\newcommand{\khoi}[1]{\textcolor{mydarkblue}{[Khoi: #1]}}
\newcommand{\chau}[1]{\textcolor{mydarkgreen}{[Chau: #1]}}
\newcommand{\truong}[1]{\textcolor{myred}{[Truong: #1]}}
\newcommand{\tuan}[1]{\textcolor{mydarkred}{[Tuan: #1]}}

\title{\Approach: Open-Vocabulary Object Detection by Linear Probing 
}


\author{%
 Chau Pham\thanks{The first two authors contribute equally.} \quad Truong Vu\footnotemark[1] \quad Khoi Nguyen \\
  VinAI Research
 \\
}

\maketitle

\begin{abstract}
    This paper addresses the challenging problem of open-vocabulary object detection (OVOD) where an object detector must identify both seen and unseen classes in test images without labeled examples of the unseen classes in training. A typical approach for OVOD is to use joint text-image embeddings of CLIP to assign box proposals to their closest text label. However, this method has a critical issue: many low-quality boxes, such as over- and under-covered-object boxes, have the same similarity score as high-quality boxes since CLIP is not trained on exact object location information. To address this issue, we propose a novel method, \Approach, that discards low-quality boxes by training a sigmoid linear classifier on pseudo labels retrieved from the top relevant region proposals to the novel text.  Experimental results on COCO affirm the superior performance of our approach over the state of the art, achieving \textbf{40.5} in $\text{AP}_{novel}$ using \textbf{ResNet50} as the backbone and without external datasets or knowing novel classes during training. Our code will be available at \url{https://github.com/VinAIResearch/LP-OVOD}.
\end{abstract}

\section{Introduction}
\label{sec:intro}

Open-Vocabulary Object Detection (OVOD) is an important and emerging computer vision problem. The task is to detect both seen and unseen classes in test images, given only bounding box annotations of seen classes in the training set. Seen classes are called base classes, while unseen classes are called novel classes and explicitly specified by their names. Novel classes are determined based on the availability of annotations for those classes in the training set. Classes present in training images without annotations are still considered novel classes.
OVOD has various applications where a detector should be capable of extending its detected categories to novel classes without human annotation such as in autonomous driving or augmented reality where new classes can appear in deployment without annotation. OVOD is also useful as an automatic labeling system in scenarios where it is impractical for annotators to exhaustively label all objects of all classes in a large dataset. 

\begin{figure}[t]
  \centering
  \includegraphics[width=.8\linewidth]{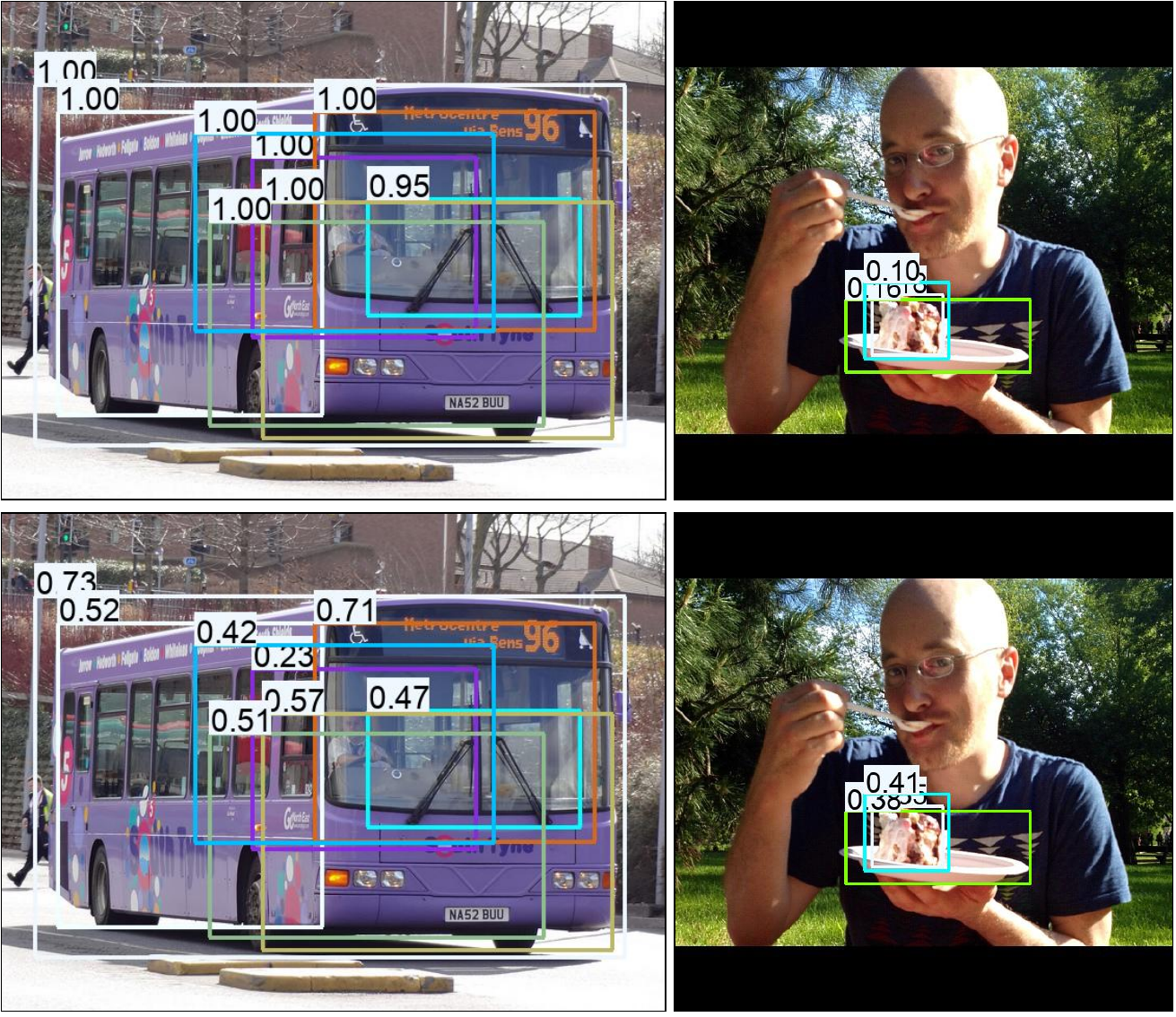}
   \caption{Comparison of box predictions for novel classes `bus' and `cake' between ViLD \cite{gu2021open} (top) and our approach (bottom). In the ViLD results, low-quality boxes have similar scores to high-quality ones, leading to high false positive (left) and false negative rates (right). Our approach significantly improves the detection performance in both cases by using classification scores instead of similarity scores as in VilD. 
   }
   \label{fig:vild_results}
\vspace{-3ex}
\end{figure}

The main challenge in OVOD is to detect novel classes without labels while maintaining good performance for base classes. To address this challenge, a pretrained visual-text embedding model, such as CLIP \cite{radford2021learning} or ALIGN \cite{jia2021scaling}, is provided as a joint text-image embedding where base and novel classes co-exist. This embedding can be used to align box proposals with their closest classes. However, the box proposals are not perfect as they are not trained on the labels of novel classes. Consequently, low-quality proposals, such as over- and under-covered-object boxes, can co-exist with high-quality ones, with the same similarity scores to their text embeddings. This is because CLIP is trained on images without object location information, leading to high false positive and false negative rates in the OVOD approaches as exemplified in Fig.~\ref{fig:vild_results}.

\begin{figure}[t]
      \centering
      \includegraphics[width=1.\linewidth]{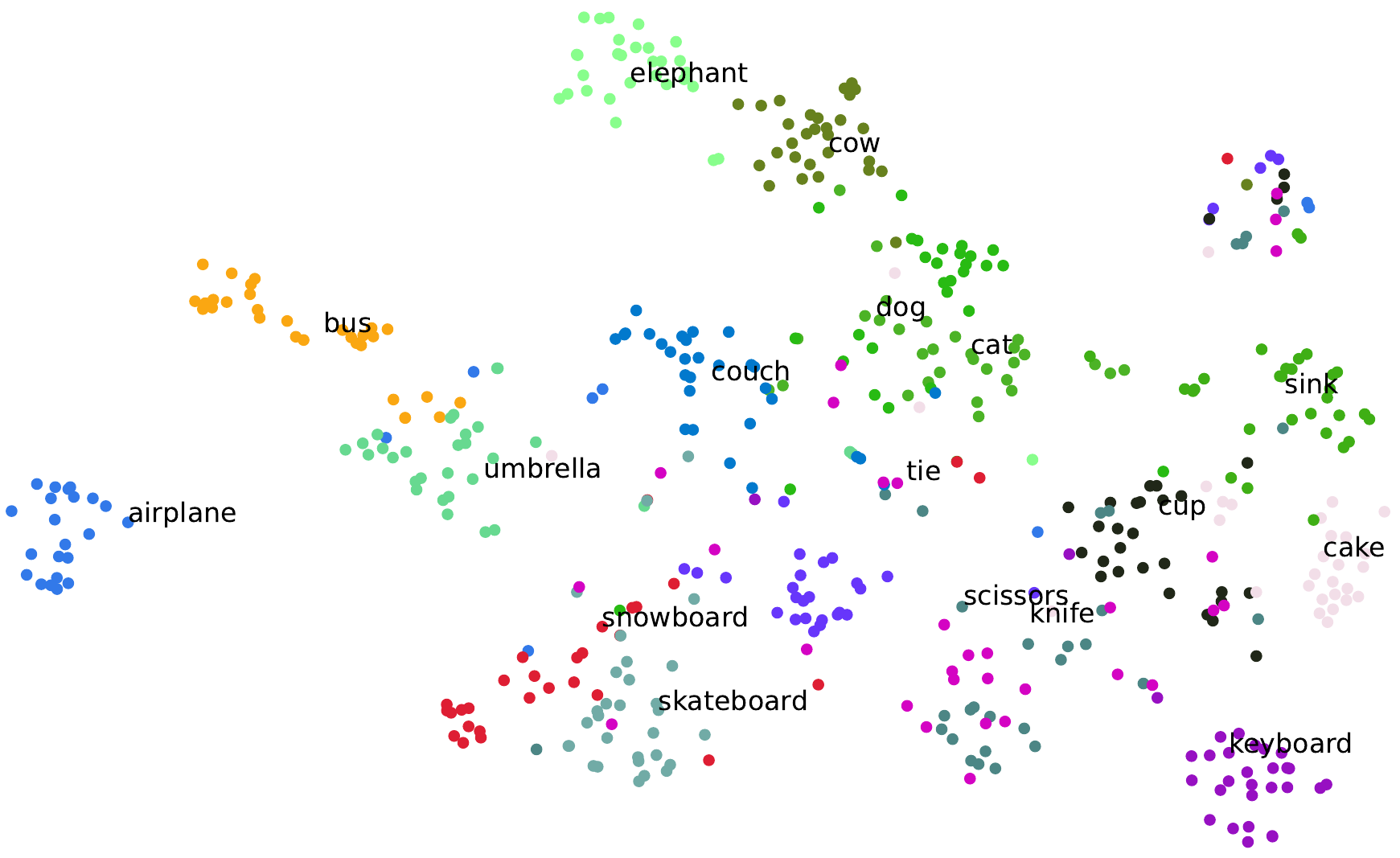}
    \caption{The feature embeddings of COCO novel classes are extracted from the penultimate layer of a Faster R-CNN pretrained on base classes. These embeddings are highly discriminative, which motivates us to learn a robust classifier from them.}
    \label{fig:branch1_embedding}
\vspace{-2.5ex}
\end{figure}

To address this limitation, we propose a novel linear probing method called \Approach~that learns a linear classifier for novel classes on top of the features extracted from the penultimate layer of a Faster R-CNN model pre-trained on base classes. These features are highly discriminative among novel classes, as shown in Fig.~\ref{fig:branch1_embedding}, despite being trained only on base classes. To obtain pseudo labels for training the linear classifier, we retrieve box candidates from the top relevant proposal boxes to the novel text. In this way, our approach leverages the presence of novel classes or similar in the training images, even in the absence of annotations. Furthermore, to facilitate quick combining with the linear classifier learned from base classes without hand-crafted calibration of the predicted scores, we propose to learn a sigmoid classifier instead of a softmax classifier for both base and novel classes since each class is predicted independently in the sigmoid classifier. Accordingly, we only need to concatenate the weights of the linear classifier of novel classes to that of base classes to enable object detection on both base and novel classes.

We demonstrate the effectiveness of our approach on two standard OVOD datasets: COCO \cite{lin2014microsoft} and LVIS \cite{gupta2019lvis}.
\Approach~significant improvement over state-of-the-art methods, without relying on external datasets or retraining the whole network whenever novel classes arrive.

In summary, the contributions of our work are as follows:

\begin{itemize}[noitemsep,topsep=0pt]
    \item A linear probing approach that leverages the highly discriminative features extracted from the penultimate layer of a pretrained Faster R-CNN on base classes to train a linear classifier for novel classes
    on the pseudo labels from retrieving the top relevant box proposals.
    \item Sigmoid classifiers for both pretraining on base classes and linear probing of novel classes to predict class scores independently, forming a unified classifier for both base and novel classes in testing.
\end{itemize}

In the following, Sec.~\ref{sec:related_work} reviews prior work; Sec.~\ref{sec:approach} specifies our approach; and Sec.~\ref{sec:experiments} presents our experimental results. Sec.~\ref{sec:conclusion} concludes with some remarks.

\begin{figure*}[t]
  \centering
  \includegraphics[width=1.\linewidth]
  {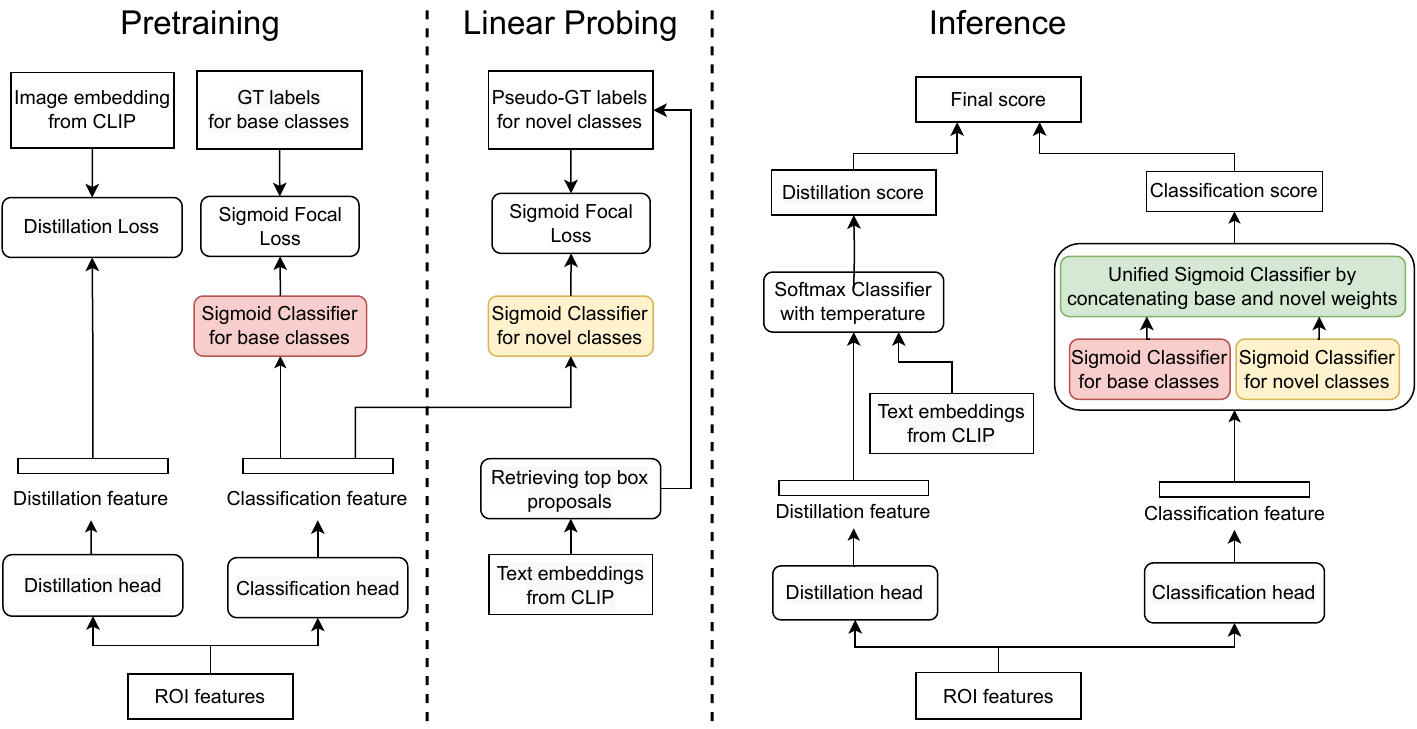}
    \caption{\textbf{Overview of our approach.} \Approach~starts from the given ROI features extracted from Faster R-CNN \cite{ren2015faster} with the same prior steps. 
   In the pretraining step \textbf{(left)}, a distillation head is added to mimic the prediction of CLIP's image encoder as in VilD \cite{gu2021open}. Furthermore, the softmax classifier is replaced with a sigmoid classifier and trained with the GT labels for the base classes. 
   In the linear probing step \textbf{(middle)}, a new sigmoid classifier with a learnable linear layer is trained on the pseudo labels of the novel classes. The pseudo labels are obtained by retrieving the top box proposals from the given novel text embedding.
   In the inference step \textbf{(right)}, we simply concatenate the weights of the two sigmoid classifiers together to form a unified sigmoid classifier for both base and novel classes where the score of each class is predicted independently. Finally, the classification scores are combined with
   the distillation score to form the final score for detection. 
   } 
   \label{fig:architecture}
\vspace{-3ex}
\end{figure*}

\section{Related Work}
\label{sec:related_work}


\noindent\textbf{Object detection} approaches aiming to localize and classify objects in images can be classified into three groups: anchor-based, anchor-free, and DETR-based detectors. Anchor-based detectors, such as Faster RCNN \cite{ren2015faster}, RetinaNet \cite{lin2017focal}, and YOLO \cite{redmon2017yolo9000}, first classify and then regress the predefined anchor boxes. In contrast, anchor-free detectors like CenterNet \cite{zhou2019objects} and FCOS  \cite{tian2019fcos} regress the bounding box extent directly without using predefined anchor boxes. DETR-based detectors \cite{carion2020end, zhu2020deformable, wang2022anchor, li2022dn, liu2022dabdetr, zhang2022dino} leverage encoder-decoder transformer architecture along with one-to-one matching loss to predict object bounding boxes in an end-to-end manner without using NMS. However, these methods are designed to work in a closed-vocabulary setting, where detectors are trained and evaluated on predefined categories and cannot detect unseen categories in testing, unlike our OVOD setting.

\myheading{Few-shot object detection (FSOD)} approaches \cite{pmlr-v119-wang20j, qiao2021defrcn, fan2020few, xiao2020few} aim to detect novel objects with a few labeled examples. 
 On the other hand, OVOD only requires the names of the novel classes instead. These two inputs are complementary since some fine-grained classes may be easier to identify through exemplars, while others may be more common and easier to identify through their names.

\myheading{Zero-shot or open-vocabulary object detection (ZSOD/ OVOD)} aims to detect unseen categories given the class name. To enable open-vocabulary learning, during training, we are provided with labeled examples of the base classes and a pretrained word embedding (such as Word2vec \cite{mikolov2013efficient}, GloVe \cite{pennington2014glove}), or vision-language models (such as CLIP \cite{radford2021learning}, ALIGN \cite{jia2021scaling}).
OVOD methods can be groupped as follows:

\textit{External-dataset-based methods} \cite{minderer2022simple,rasheed2022bridging,feng2022promptdet,zhou2022detecting,gao2021towards,zhong2022regionclip,zareian2021open,Huynh:CVPR22, Bravo2022locov,VLDet} utilize huge external datasets, including image-caption pairs or image-level labeled annotations, to improve the pretrained vision-language model or detectors to recognize more classes, including the novel ones. Thus, these methods have an advantage over those that do not.

\textit{Novel-class-aware methods} including OV-DETR \cite{zang2022open}, VL-PLM \cite{zhao2022exploiting} assume that novel categories are known during training. These methods retrieve large-scale region proposals of novel classes based on the joint text-image embedding of CLIP \cite{radford2021learning} as pseudo-GT labels, which are jointly trained with GT-labeled examples of base classes. As a result, these methods need to regenerate the pseudo labels and retrain the detectors whenever new classes arrive.

\textit{Novel-class-unaware methods} \cite{gu2021open, du2022learning,kuo2022f} follow the same setting as ours. ViLD \cite{gu2021open} uses knowledge distillation from CLIP visual features to learn the embedding for unseen categories. DetPro \cite{du2022learning} proposes a learnable-text prompt instead of a fixed-text prompt. 
F-VLM \cite{kuo2022f} utilizes a pretrained CLIP's image encoder as a backbone to retain the locality-sensitive features necessary for detection.

However, these methods attempt to align the text embedding with the feature embedding of each proposal to predict its class. In contrast, our method approaches a different way that learns a linear classifier for novel classes using features extracted from a Faster R-CNN pretrained on base classes.

\section{Our Approach}
\label{sec:approach}

\myheading{Problem statement:}
During training, we are provided with a large set of annotated examples of base classes $C_B$, i.e., bounding boxes $b_i$ and their categories $c_i \in C_B$. In testing, given the names of novel classes $C_N$, our goal is to detect objects of both base and novel classes, i.e., ${\hat{c}_i, \hat{b}_i}$, where $\hat{c}_i \in C_B \cup C_N$ for test images. To facilitate learning, a pretrained CLIP \cite{radford2021learning} is provided as the joint image-text embedding of both base and novel classes.

\myheading{Our scope:} Our approach strictly assumes that we do not know novel classes during training, as we cannot anticipate the classes that an open-vocabulary detector (OVD) will encounter in practical use. 
Additionally, to ensure a fair comparison, we utilize only the images and annotations provided by each benchmark without any external datasets, such as image-caption or image-level label datasets.

Fig.~\ref{fig:architecture} illustrates our approach, which is based on Faster R-CNN \cite{ren2015faster}. We adopt the same backbone, region proposal network (RPN), and box regression modules, and refer readers to \cite{ren2015faster} for details. However, we make two modifications: replacing the softmax classifier with a sigmoid classifier and adding a distillation head as in ViLD \cite{gu2021open}.
For novel classes, we extract features from the top relevant proposals to the novel text embedding as pseudo labels for training a sigmoid classifier of the novel classes. In testing, we concatenate the weights of the two sigmoid classifiers to form a unified sigmoid classifier for object detection.

\subsection{Pretraining on Base Classes} 
\label{sec:base_training}

As motivated in the introduction, to facilitate the fast learning of novel classes when they arrive in testing, we propose to replace the softmax classifier of Faster R-CNN \cite{ren2015faster} with a sigmoid classifier to pretrain on base classes. In this way, instead of classifying among different categories and a background class, we predict the presence or absence of a category in an image. In other words, the embeddings of new classes are distributed diversely far from those of the base classes rather than grouping together into a `background' class as shown in Fig.~\ref{fig:branch1_embedding}. Also,
such a classifier predicts each category independently so that when new classes arrive, we can incrementally concatenate the weights of the newly trained classifier to that of the base classes to form a new unified classifier that can readily work on both base and new classes without any retraining or temperature tuning. 




%
Concretely, the ROI features for proposals $\tilde{b}_i$ are extracted from the backbone and forwarded to the classification head and distillation head to obtain classification feature $f^{\text{cls}}_i$ and distillation feature $f^{\text{dis}}_i$, respectively. We then jointly train the new sigmoid classifier and the distillation head. The sigmoid classifier is supervised by the ground-truth labels $c_i$ of the base classes using sigmoid focal loss \cite{lin2017focal}. Meanwhile, the distillation head is supervised by the CLIP image embedding $e^{\text{image}}_i$, which is obtained from CLIP's image encoder using cropped images from the proposal $\tilde{b}_i$. The distillation head is trained using the L1 loss. In particular,
\begin{align}
    \mathcal{L}_{\text{cls}}^{\text{Base}} &= \sum_{i} \textbf{Focal loss}(\text{Sigmoid}(f_i^\text{cls}; W_B), c_i), \\
    \mathcal{L}_{\text{dis}} &= \sum_i\|{f^{\text{dis}}_i - e^{\text{image}}_i}\|_1,
\end{align}
where $W_B$ are the weights of the base classes.

\begin{figure}[t]
  \centering
  \includegraphics[width=1.0\linewidth, trim={0 9cm 0 0},clip]{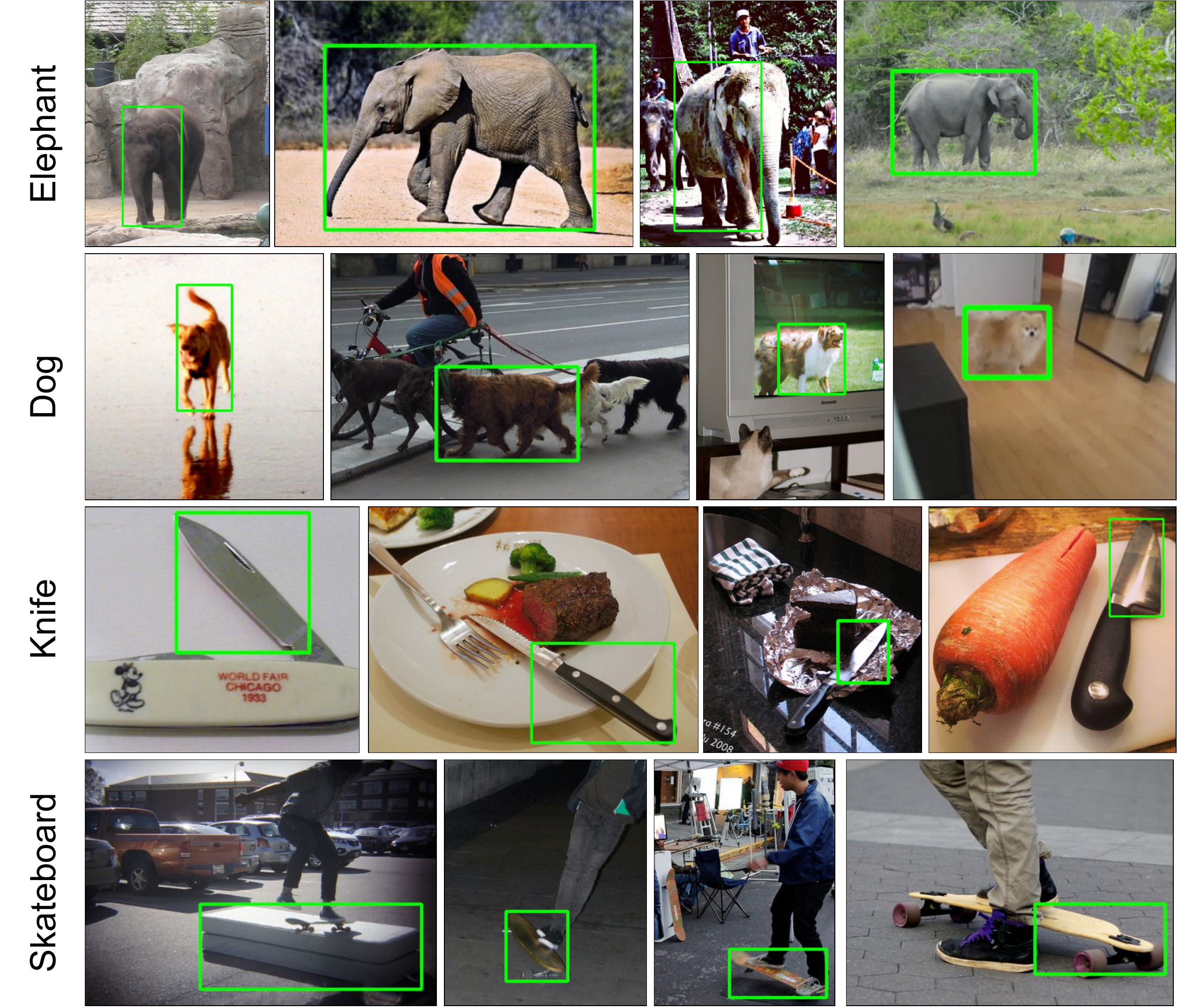}
   \caption{Top-4 box proposal retrievals from CLIP's embeddings of four novel classes: `elephant', `dog', and `knife'.
   The quality is good enough to be used as pseudo labels for training a few-shot classifier on novel classes.
   }
   \label{fig:retrieval}
\vspace{-3.ex}
\end{figure}

\subsection{Linear Probing on Novel Classes}
\label{sec:fewshot_training}

As illustrated in Fig.~\ref{fig:vild_results}, low-quality boxes usually have the same similarity score to the novel text embeddings as the high-quality ones do, resulting in high false positive and false negative rates. 
%
Therefore, we need to have better positive/negative proposals for training a sigmoid classifier to discard these low-quality proposals. 

To this end, first, the top relevant proposals of each novel class are retrieved and served as pseudo-GT labels $\tilde{c}_i$. Specifically, we extract all image embeddings $e_i^{\text{image}}$ of all proposals $\tilde{b}_i$ having the objectness score $o_i$ larger than $\tau$ in the training set. For each novel category with text embedding $e^{\text{text}}_c$ where $c\in C_N$, we retrieve the top $K$ closest proposals 
in order to form a set $\mathcal{P}=\{(\tilde{b}_i, \tilde{c}_i)\}_{i=1..K\times C_N}$
using cosine similarity $\text{cos}({e^{\text{text}}_c}, e_i^{\text{image}})$. We visualize the examples of top-4 retrieved proposal for four novel classes in Fig.~\ref{fig:retrieval}.
To speed up the retrieval process, we resort the Faiss \cite{johnson2019billion} tool.
Then, we leverage the sampling mechanism of Faster R-RCNN to sample positive/negative proposals where the positives $\mathcal{P}^+=\{(\tilde{b}_i, \tilde{c}_i)\}, \tilde{c}_i \in N_c$ are the ones having IoU $> 0.5$ with the pseudo-GT boxes $\mathcal{P}$ while the rest are the negatives $\mathcal{P}^-=\{(\tilde{b}_i, 0)\}$. 
When novel classes arrive, 
a new sigmoid classifier $W_N$ is added on top of the pretrained classification feature $f^{\text{cls}}_i$. 
The sigmoid classifier for novel classes is trained as follows:
\begin{align}
    \mathcal{L}_{\text{cls}}^{\text{Novel}} &= \sum_{i=1}^{|\mathcal{P}^+ \cup \mathcal{P}^-|} \textbf{Focal loss}(\text{Sigmoid}(f_i^{\text{cls}}; W_N), c_i),
\end{align}
where $W_N$ are the weights of the novel classes.

Notably, our approach is fast, i.e., 5 minutes on COCO, because we only retrieve the top proposals. This differs from OV-DETR \cite{zang2022open} and VL-PLM \cite{zhao2022exploiting}, which extract pseudo labels for new classes from the entire training set and jointly train them with the ground truth labels for the base classes.

\subsection{Inference on Both Base and Novel Classes}
\label{sec:inference}
Given a proposal box $\tilde{b}_i$ with classification feature $f_i^{\text{cls}}$ and distillation feature $f_i^{\text{dis}}$, the inference on both base and novel classes is visualized in the right of Fig.~\ref{fig:architecture}.

For the classification head, we concatenate the weights of the sigmoid classifiers learned on the base and novel classes to form a unified classifier with weight $W = [W_B; W_N]$. The classification score $s^{\text{cls}}_i$ is calculated as:
\begin{equation}
    s^{\text{cls}}_i = \text{Sigmoid}(f_i^{\text{cls}}; W) \in [0, 1]^{|C_B|+|C_N|}. 
\end{equation}
 
For the distillation head, we compute the distillation score $s^{\text{dis}}_i$ as the softmax score of the cosine similarity between the distillation features $f^{\text{dis}}_i$ and text embeddings $e^{\text{text}}_c$ with temperature $\kappa$ as:
\begin{equation}
    s^{\text{dis}}_i = \text{Softmax}_c \left(\frac{\cos(f_i^{\text{dis}}, e_c^{\text{text}})}{ \kappa} \right) \in [0, 1]^{|C_B|+|C_N|}.
\end{equation}

Finally, the final score for prediction of each proposal $\tilde{b}_i$ with objectness score $o_i$ is computed as:
\begin{align}
    s_i &=o_i \cdot
    \begin{cases}
    s^{\text{cls}}_i \text{ for base classes} \\
    (s^{\text{cls}}_i)^{\beta}(s^{\text{dis}}_i)^{1-\beta} \text{ for novel classes}
    \end{cases} 
    \label{eq:final_score}
\end{align}
where $\beta$ are coefficient hyper-parameter for novel classes.


\section{Experimental Results}
\label{sec:experiments}

\begin{table*}[t]
\centering
{
\small
\begin{tabular}{lccc>{\color{gray!70}}ccc>{\color{gray!70}}c>{\color{gray!70}}cc}
\toprule
\multirow{2}{*}{\textbf{Method}} & \multirow{2}{*}{\textbf{Venue}} & \multirow{2}{*}{\textbf{Training source}} & \multicolumn{3}{c}{\textbf{Box AP on COCO} } & \multicolumn{4}{c}{\textbf{Mask AP on LVIS} } \\
& & & \textbf{$\text{AP}_{novel}$} &  \textbf{$\text{AP}_{base}$} & \textbf{AP} & \textbf{$\text{AP}_r$} & \textbf{$\text{AP}_f$} & \textbf{$\text{AP}_c$} & \textbf{AP}\\
\midrule

\rowcolor{gray!30} OVR-CNN \cite{zareian2021open} & CVPR 21 &    & 22.8 & 46.0 & 39.9 & - & - & - & - \\
\rowcolor{gray!30} XPM \cite{Huynh:CVPR22} & CVPR 22 & & 27.0 & 46.3 & 41.3 & - & - & - & -\\
\rowcolor{gray!30} RegionCLIP \cite{zhong2022regionclip} & CVPR 22 & & 31.4 & 57.1 & 50.4 
& 17.1 & 27.4 & 34.0 & 28.2 \\
\rowcolor{gray!30} PromptDet \cite{feng2022promptdet} & ECCV 22 & & 26.6 &59.1 & 50.6 
& 19.0 & 18.5 & 25.8 & 21.4\\
\rowcolor{gray!30} Detic \cite{zhou2022detecting} & ECCV 22 & & 27.8 & 47.1 & 42.1 & 
17.8 & 26.3 & 31.6 & 26.8 \\
\rowcolor{gray!30} PB-OVD \cite{gao2021towards} & ECCV 22 & & 30.8 & 46.1 & 30.1 & - & - & - & -\\
\rowcolor{gray!30} OWL-ViT \cite{minderer2022simple} & ECCV 22 & & 41.8 & 49.1 & 47.2 & 16.9 & - & - & 19.3 \\
\rowcolor{gray!30} VLDet \cite{VLDet} & ICLR 23 & \multirow{-8}{*}{\makecell{image captions in $C_B \cup C_N$\\instance-level labels in $C_B$ \\ (use external datasets)}}  & 32.0 & 50.6 & 45.8 & 21.7 & 29.8 & 34.3 & 30.1 \\
\midrule
\rowcolor{gray!30} OV-DETR \cite{zang2022open} & ECCV 22 & &29.4 & 61.0 &52.7
& 17.4 & 25.0 & 32.5 & 26.6\\
\rowcolor{gray!30} VL-PLM \cite{zhao2022exploiting} & ECCV 22 & \multirow{-2}{*}{\makecell{instance-level labels in $C_B$ \\ known novel classes during training}} & 34.4 &60.2 & 53.5 &  17.2	& 23.7 & 35.1 & 27.0 \\
\midrule
ZSD \cite{bansal2018zero}  & ECCV 18      & \multirow{3}{*}{\makecell{instance-level labels in $C_B$ \\(zero-shot object detection)}}  & 0.31 & 29.2 & 24.9 & - & - & - & - \\
PL \cite{rahman2020improved} & AAAI 20  &       & 4.12 & 35.9 & 27.9 & - & - & - & - \\
DELO \cite{zhu2020don} & CVPR 20      &       & 3.41 & 13.8 & 11.1 & - & - & - & -\\

\midrule
ViLD \cite{gu2021open} & ICLR 22 & \multirow{5}{*}{\makecell{instance-level labels in $C_B$ \\ unknown novel classes during training}}  & 27.6 & 59.5 & \underline{51.2} 
& 16.6 & 24.6 & 30.3 & 25.5 \\
RegionCLIP$^{\dagger}$ \cite{zhong2022regionclip} & CVPR 22 &   & 14.2 & 52.8 & 42.7 & - & - & - & -\\
DetPro$^{\ddagger}$ \cite{du2022learning} & CVPR 22 & &19.8 & 60.2 & 49.6 & \textbf{19.8} & 25.6	& 28.9 & \underline{25.9} \\
F-VLM \cite{kuo2022f} & ICLR 23 & &  \underline{28.0} &43.7 &39.6 & 18.6 & - & - & 24.2 \\
\Approach~(ours) & - & & \textbf{40.5} & 60.5 & \textbf{55.2} & \underline{19.3} & 26.1 & 29.4 & \textbf{26.2} \\
\bottomrule
\end{tabular}
}
\caption{
\textbf{Performance on COCO and LVIS.} `-' denotes numbers that are not reported. $^{\dagger}$ denotes another version of RegionCLIP using only the COCO object detection dataset for training. $^{\ddagger}$ denotes our re-run of the provided DetPro source code on COCO without the transferring from LVIS. Methods in \colorbox{gray!30}{faded rows} are for reference only, not a direct comparison to ours. Best results are in \textcolor{black}{\textbf{bold}} and the second best are in \textcolor{black}{\underline{underlined}
}.
}
\label{tab:results}
\vspace{-2.5ex}
\end{table*}

\myheading{Datasets:}
We conduct our experiments using the OVOD versions called OV-COCO \cite{bansal2018zero} and OV-LVIS \cite{gu2021open} of  two public datasets: COCO \cite{lin2014microsoft} and LVIS \cite{gupta2019lvis}. The OV-COCO dataset comprises 118,000 images with 48 base categories and 17 novel categories.
OV-LVIS \cite{gupta2019lvis} shares the image set with OV-COCO.
Its categories are divided into `frequent', `common', and `rare' groups based on their occurrences, representing the long-tailed distributions of 1,203 categories. We treated the `frequent' and `common' groups of 866 categories as the base classes while considering the rare' group of 337 categories as the novel classes. 


\myheading{Evaluation metrics:} Consistent with the standard OVOD evaluation protocol \cite{gu2021open, zhong2022regionclip}, we report the box Average Precision (AP) with an IoU threshold of 0.5 for object detection on the COCO dataset, i.e. $\text{AP}_{novel}$ for novel classes, $\text{AP}_{base}$ for base classes, and AP for all classes. For instance segmentation on the LVIS dataset, we report the mask AP, which is the average AP over IoU thresholds ranging from 0.5 to 0.95, i.e., $\text{AP}_r, \text{AP}_{f}$, $\text{AP}_c$, and AP for `rare', `frequent', `common', and all classes, respectively. 

\myheading{Implementation details:}
In our implementation, we use the Faster R-CNN detector \cite{ren2015faster} for COCO and the Mask-RCNN detector \cite{he2017mask} for LVIS, both with the ResNet50 \cite{he2016deep} backbone. The ResNet50 backbone is initialized with the self-supervised pre-trained SoCo \cite{wei2021aligning}. We use multi-scale training with different image sizes while maintaining the aspect ratio for data augmentation. We employ OLN \cite{kim2022learning} as the object proposal network. 
For training on base classes, we use the SGD optimizer with an initial learning rate of 0.02 and an image batch size of 16. We adopt the 20-epoch schedule from MMDetection \cite{mmdetection}, where the learning rate is decreased by a factor of 10 after the 16th and 19th epochs, and apply a linear warm-up learning rate for the first 500 iterations.
For quick adapting to novel classes, we set the objectness score threshold to $\tau=0.6$ to filter proposals before retrieval. We train the novel weights $W_N$ for 12 epochs using the SGD optimizer with an initial learning rate of 0.01 and decreasing the learning rate by a factor of 10 after the 8th and 11th epochs. In testing, we use a temperature of $\kappa=0.01$ for the distillation head.

\begin{figure*}[t]
  \centering
  \includegraphics[width=.93\linewidth]{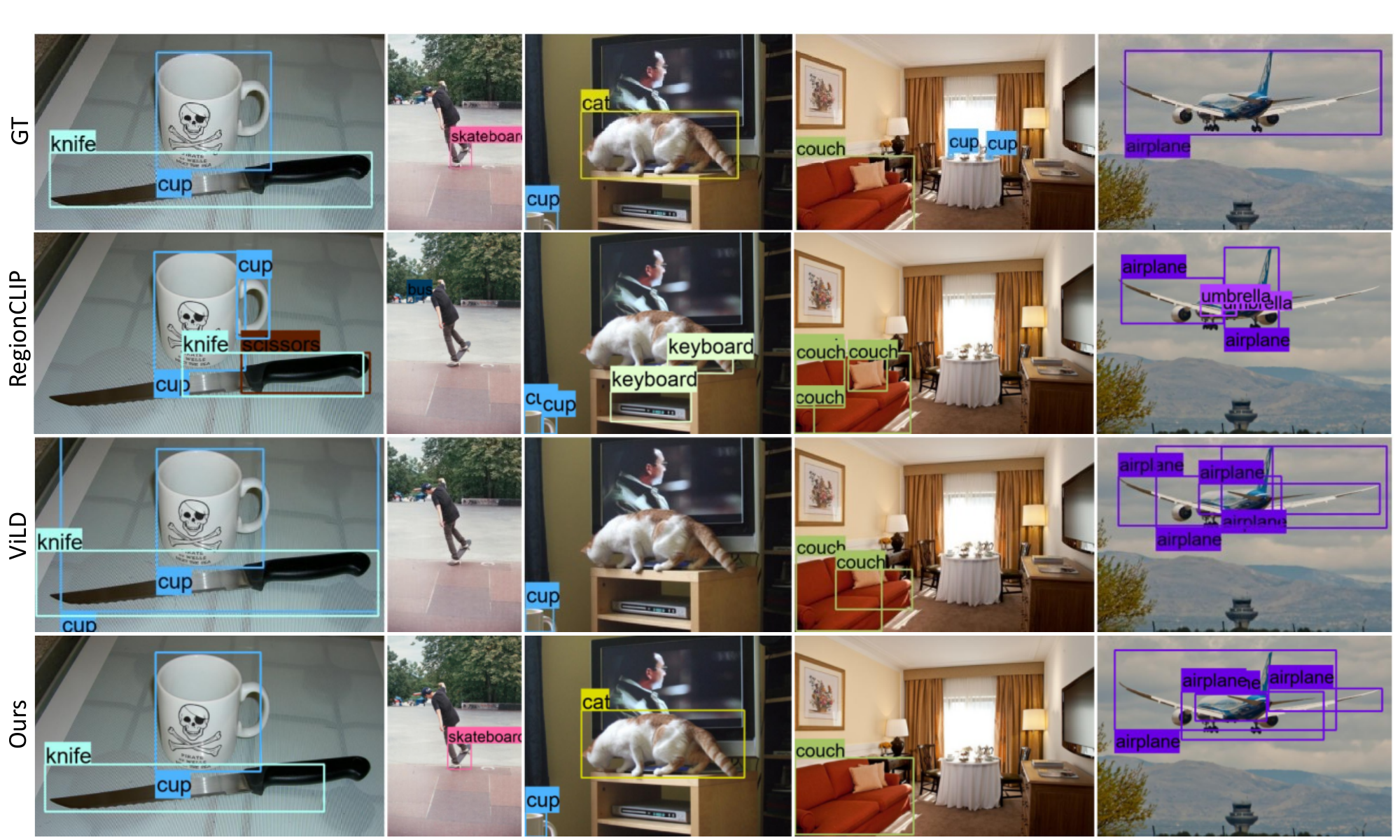}
   \caption{\textbf{Qualitative comparison of different approaches on COCO's novel classes.} The first four columns show our superior performance while the last one shows a failure case where all of them cannot generate boxes for the airplane due to its rare aspect ratio.}
   \label{fig:qualitative_res}
\vspace{-2ex}
\end{figure*}


\subsection{Comparison with State-of-the-art Approaches}

\myheading{Results on COCO} are shown in 
Tab.~\ref{tab:results} and Fig.~\ref{fig:qualitative_res}. In Tab.\ref{tab:results}, we compare our approach to various methods, including ZSOD, external-dataset-based, novel-class-aware, and novel-class-unaware methods. Our approach significantly outperforms the second-best method on COCO with a significant margin of +11.5 in $\text{AP}_{novel}$, while maintaining good performance on base classes. In Fig.~\ref{fig:qualitative_res}, our approach achieves superior performance, while RegionCLIP incorrectly classifies foreground instances as background and ViLD generates redundant predictions. The last column shows a failure case, where all methods struggle to generate accurate boxes for the airplane due to its aspect ratio being significantly different from the base classes. Therefore, these results demonstrate the effectiveness of our approach without relying on any external datasets or known novel classes during training.

\begin{figure*}[t!]
    \centering
    \includegraphics[width=0.93\linewidth]{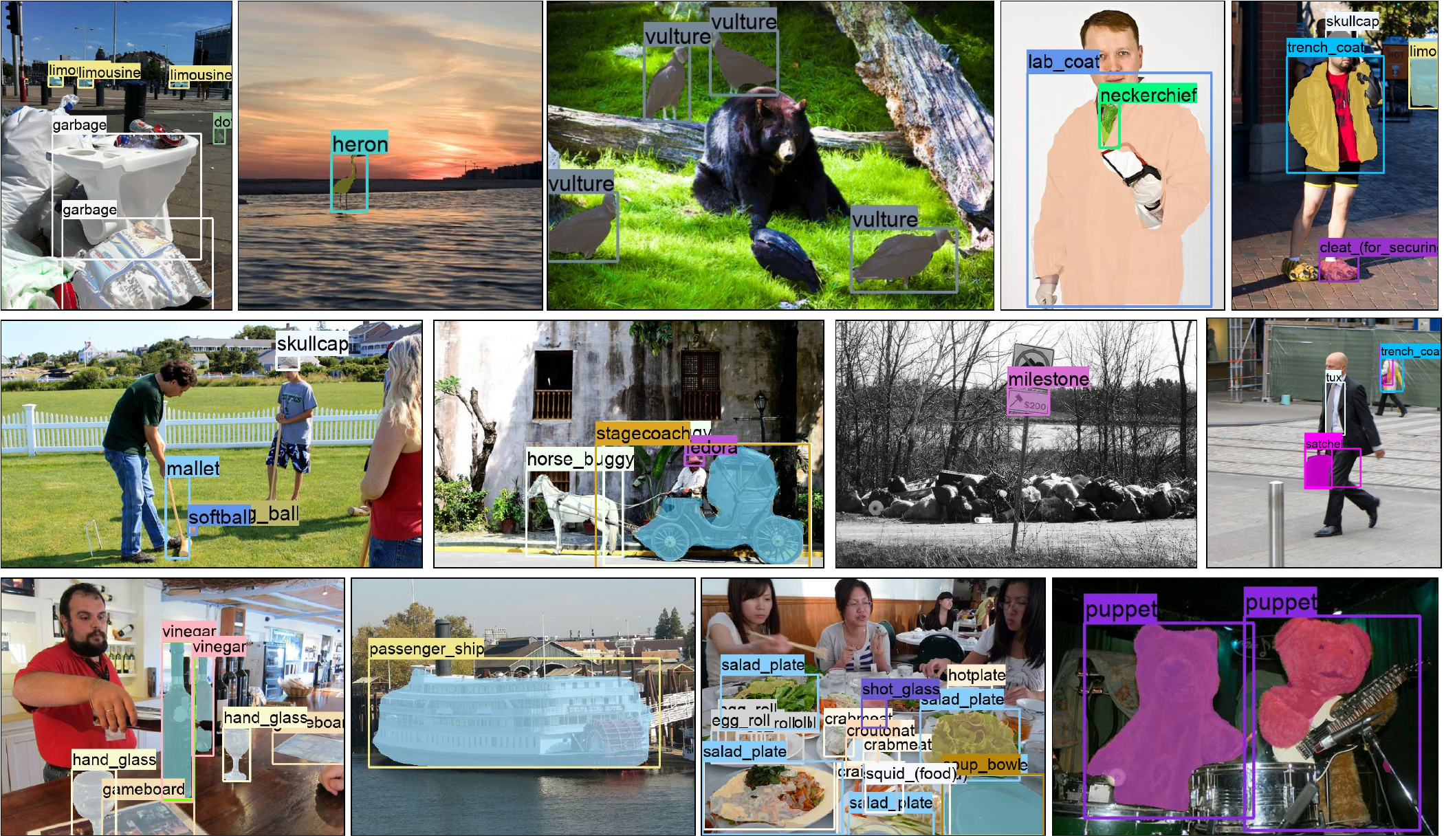}
    \caption{\textbf{Qualitative results of novel classes on the LVIS dataset \cite{gupta2019lvis}}. Our approach can successfully detect some novel classes including ``lab coat", ``mallet", and ``hand glass". However, due to the rarity of some novel classes in training, our method retrieves the proposals of close-meaning classes instead, i.e.,  ``tie" vs ``neckerchief", leading to the wrong prediction in testing.}
    \label{fig:lvis_res}
    \vspace{-2ex}
\end{figure*}

%




\myheading{Results on LVIS} are shown in Tab.~\ref{tab:results} and Fig.~\ref{fig:lvis_res}. As can be seen, we obtain comparable results with DetPro \cite{du2022learning} and the improvement is less significant than that in COCO.
That is due to two main reasons. 
\textit{First}, the semantic difference between base and novel classes in COCO is relatively high owing to the smaller number of classes while the difference in LVIS is lower since classes are more fine-grained, giving rise to easier transfer of the learned embedding in base classes to novel classes in LVIS. Hence, novel text embeddings are readily matched with the predicted feature in testing. 
\textit{Second}, our method is mainly based on the assumption that novel classes exist even though they are not annotated in training images. As a result, the performance of the few-shot learner mostly depends on the quality of the retrieved proposal given novel classes' names. In LVIS, the distribution of classes is long-tail, especially for rare classes which are tested as novel classes. All of them appear less than 10 times in the training set. It is very challenging for our approach to retrieve relevant proposals. Fortunately, even though our method cannot retrieve the exact proposals for each novel class, it can retrieve the close-meaning proposals such as `neckerchief' vs. `tie', `puppet' vs. `doll', and `elephant' vs. `mammoth'. Thus, our method performs comparably with prior work in LVIS.

\begin{table}
    \setlength{\tabcolsep}{3pt}
    \centering
    \begin{tabular}{c c c}
         \toprule
         & \textbf{$\text{AP}_{novel}$ on COCO} & \textbf{$\text{AP}_{r}$ on LVIS} \\
         \midrule
         Ours + RPN \cite{ren2015faster} & 37.2 & 19.3 \\
         Ours + OLN \cite{kim2022learning} & \textbf{40.5} & 19.3 \\
         \bottomrule
    \end{tabular}
    \caption{The effectiveness of RPN \cite{ren2015faster} and OLN \cite{kim2022learning}.}
    \label{tab:oln}
\vspace{-2.ex}
\end{table}

\subsection{Ablation Study}
In this section, we conduct ablation studies on the COCO dataset on various aspects to analyze our approach.

\myheading{Impact of different proposal networks.} Tab.~\ref{tab:oln} presents the results of our approach using RPN \cite{ren2015faster} and OLN \cite{kim2022learning} proposals. OLN is a SOTA object proposal network in the open-world setting. On COCO, the quality of the OLN proposals is higher than that of RPN with the same supervision in training, as evidenced by an improvement of +3.3 in $\text{AP}_{novel}$. This is because OLN is more robust to object sizes and aspect ratios by replacing foreground/background classification with centerness and IoU score predictions. However, when the number of base classes increases, as in the case of LVIS, these predictions become less effective since the base classes can cover a wider range of object sizes and aspect ratios of the novel classes.

\begin{table}[t]
    \setlength{\tabcolsep}{3pt}
    \centering
    \begin{tabular}{c c c c c}
         \toprule
          \textbf{Retrieval} & \textbf{Sigmoid} & \textbf{$\text{AP}_{novel}$} & \textbf{$\text{AP}_{base}$} & \textbf{AP} \\
         \midrule
         & & 27.6 & 61.2 & 52.4 \\
         \checkmark  & & 33.2 & \textbf{61.2} &53.9 \\
         \checkmark &\checkmark &\textbf{40.5} & 60.5 &\textbf{55.2} \\
         \bottomrule
          
    \end{tabular}
    \caption{Ablation study on the contribution of each component. \textbf{Retrieval}: retrieving top boxes as pseudo labels for novel classes. \textbf{Sigmoid}: replace softmax with sigmoid classifier.}
    \label{tab:component_abl}
\vspace{-2ex}
\end{table}

\myheading{Ablation study on each component's contribution} is summarized in Tab. \ref{tab:component_abl}.
Our baseline is ViLD with OLN proposals.
By using retrieval of top boxes as the pseudo labels for novel classes, the performance improves significantly by +5.6 in $\text{AP}_{novel}$ compared to the baseline, while keeping the performance of base classes intact. Moreover, combining the sigmoid classifier and the pseudo-labeling strategy results in the best performance of 40.5 in $\text{AP}_{novel}$.

          

\begin{table}[t]
    \centering
    \begin{tabular}{c c c c}
    \toprule
     \textbf{Features} & \textbf{$\text{AP}_{novel}$} & \textbf{$\text{AP}_{base}$} & \textbf{AP} \\
    \midrule
    Classification &\textbf{35.9} &60.5 &\textbf{54.1} \\
    Distillation &19.7 &60.5 &49.8 \\
    \bottomrule
    \end{tabular}
    \caption{Types of features to learn the sigmoid linear classifier.}
    \label{tab:feat_repre}
\end{table}

\myheading{Study on features to learn the sigmoid classifier.}
%
To quantitatively show that the classification features of Faster R-CNN pre-trained on base classes are superior, we train a sigmoid classifier on top of the classification feature $f^{\text{cls}}_i$ and the distillation feature $f^{\text{dis}}_i$, which is trained to distill the CLIP's image embedding. The results are presented in Tab.~\ref{tab:feat_repre}. The feature of the classification head yields 35.9 in $\text{AP}_{novel}$, greatly outperforming that of the distillation head.


\begin{table}[t]
    \centering
    \begin{tabular}{c c c c c c c}
         \toprule
         \textbf{\# proposals} & 5 & 10 & 20 & 50 & 100 & 200 \\
         \midrule
         \textbf{$\text{AP}_{novel}$} & 30.5 & 34.8 & 38.3 & 40.3 & \textbf{40.5} & 39.6 \\
         \bottomrule
    \end{tabular}
    \caption{Ablation on $\#$ retrieved proposals per novel class.}
    \label{tab:num_proposals}
\vspace{-2ex}
\end{table}

\myheading{Number of retrieved proposals per novel class.}
Tab.~\ref{tab:num_proposals} presents the performance of our approach for different numbers of proposals $K$ per novel class in Sec.~\ref{sec:fewshot_training}. The performance improves as the value of $K$ increases and saturates at K=100. We speculate that a higher number of proposals provides more diverse examples for training whereas too many proposals increase the likelihood of including noisy boxes, resulting in suboptimal performance. Moreover, too many proposals can slow down the retrieval and few-step training of the linear classifier for novel classes.




\begin{table}
    \centering
    \begin{tabular}{c c c c c}
    \toprule
         \textbf{$\beta$} & 0.9 & 0.8 & 0.7 & 0.6 \\
         \midrule
         \textbf{$\text{AP}_{novel}$} & 40.2 & \textbf{40.5} & 39.7 & 38.5 \\
         \bottomrule
    \end{tabular}
    \caption{Study on the coefficient of novel classes $\beta$.}
    \label{tab:abl_study_beta}
\end{table}

\begin{table}
    \centering
    \begin{tabular}{c c c c}
         \toprule
         \textbf{Objectness} & \textbf{$\text{AP}_{novel}$} & \textbf{$\text{AP}_{base}$} & \textbf{AP} \\
         \midrule
          & 34.6 & \textbf{61.4} & 54.4 \\
         \checkmark & \textbf{40.5} & 60.5 & \textbf{55.2} \\
         
         \bottomrule
    \end{tabular}
    \caption{The importance of objectness score $o_i$ in Eq.~\eqref{eq:final_score}.}
    \label{tab:objectness}
\end{table}

\myheading{Study on the coefficient of novel classes $\beta$} is summarized in 
Tab.~\ref{tab:abl_study_beta}. ViLD uses $\beta = 1/3$, indicating that the distillation head's novel scores have more impact on the final prediction than the classification head's scores. However, in our case, we achieve the best performance when using $\beta = 0.8$, implying that the classification score has a greater contribution than the distillation score to the final score.

\myheading{The importance of the objectness score in Eq.~\eqref{eq:final_score}.}
We compare the performance of our model with and without multiplication of the objectness score $o_i$. The object detector's objectness score provides an indication of the presence of an object in an image. Hence, multiplying the final score by the objectness score can mitigate false positive and false negative detections. In Tab.~\ref{tab:objectness}, we observe a performance gain of +5.9 in $\text{AP}_{novel}$ with the multiplication of the objectness score compared to the model without it.

\begin{figure}[t]
  \centering  \includegraphics[width=.65\linewidth,height=.17\textheight]{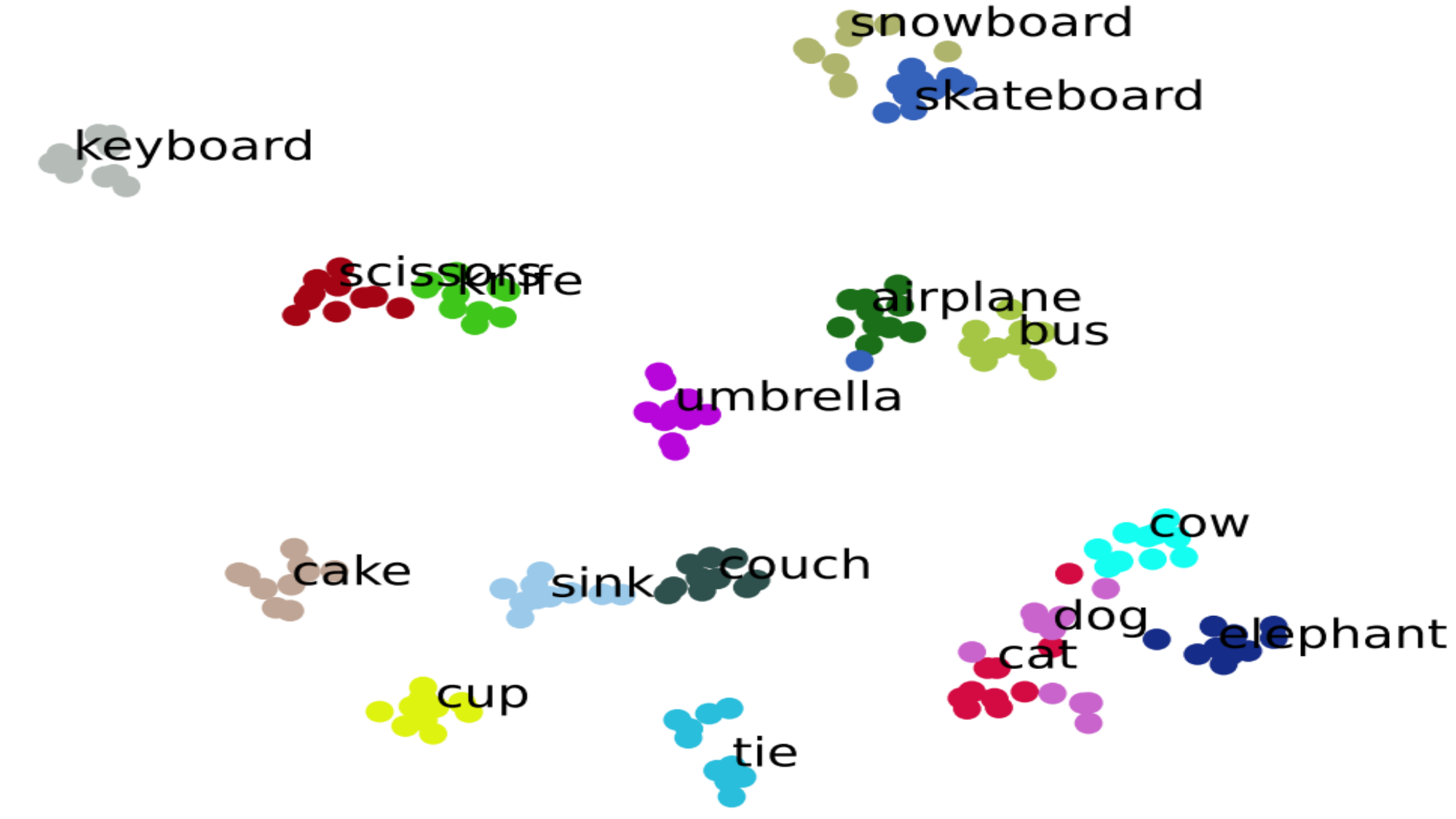}
   \caption{The CLIP's image embeddings of top retrieved boxes.
   }
   \label{fig:embedding}
\end{figure}

\myheading{Reason to choose top retrieved boxes as pseudo labels.}
Unlike the CLIP features of random proposals, the top-retrieved boxes are distinct as visualized in Fig.~\ref{fig:embedding}. Therefore, these top-retrieved boxes are good candidates for training the sigmoid classifier for novel classes.

\subsection{Transfer from LVIS to Objects365 and VOC}

We evaluate the transfer learning performance of our approach on Objects365 \cite{shao2019objects365} and PASCAL VOC \cite{everingham2010pascal} datasets, following the protocol in \cite{gu2021open,du2022learning}. We use a pretrained model on the LVIS dataset, which includes the `frequent' and `common' classes, and evaluate its performance on the validation sets of Objects365 and PASCAL VOC, consisting of 365 and 20 classes, respectively. For Objects365, we use part V1 of the newly released Objects365 V2 dataset, 
consisting of 30,310 images and over 1.2M bounding boxes. 
For PASCAL VOC, we retrieve the top $K=10$ proposals per novel class for Objects365 and the top $K=50$ proposals for PASCAL VOC and set $\beta=0.6$. Results are reported in Tab.~\ref{tab:compare365_voc}. Our approach outperforms ViLD \cite{gu2021open} and DetPro \cite{du2022learning} with a substantial margin of approximately +1.5 in APs, demonstrating the effectiveness of our approach in various transfer learning settings beyond COCO and LVIS.

\begin{table}[t]
    \centering
    \begin{tabular}{l c c c c c}
         \toprule
         &\multicolumn{3}{c}{\textbf{Objects365}} &\multicolumn{2}{c}{\textbf{PASCAL VOC}} \\
         \textbf{Method} & \textbf{AP} &\textbf{AP50} &\textbf{AP75} &\textbf{AP50} &\textbf{AP75} \\
         \midrule
         ViLD$^{\dagger}$ \cite{gu2021open} & 10.2 & 16.2 & 10.9 & 72.2 & 56.7 \\
         DetPro \cite{du2022learning} & 10.9 & 17.3 & 11.5 & 74.6 & 57.9 \\
         \midrule
         Ours & \textbf{12.6} & \textbf{18.9} & \textbf{13.1} & \textbf{76.0} & \textbf{59.4} \\
         \bottomrule
    \end{tabular}
    \caption{Transfer from LVIS to Objects365 and PASCAL VOC. $^{\dagger}$denotes the re-implementation of ViLD in the DetPro repository.}
    \label{tab:compare365_voc}
\end{table}

\section{Discussion and Conclusion}
\label{sec:conclusion}

\myheading{Limitations:} As shown in Tab.~\ref{tab:results}, the performance of novel classes is still lagging behind that of base classes, with a gap of 20 points in Box AP on the COCO dataset. One of the main reasons for this is that we did not fine-tune or improve the box regression for novel classes, as we only focused on the classification head. This is due to the lack of box annotations for novel classes, which is a common issue in OVOD. Additionally, CLIP's visual embeddings are not highly sensitive to the precise box location but only require that the box contains the object or important parts of the object. As a result, there is limited information available for improving the bounding boxes based solely on CLIP. Therefore, further research on improving box regression would be an interesting direction for OVOD.

\myheading{Conclusion:}
In this work, we have introduced a simple yet effective approach for OVOD with two contributions. Firstly, we propose a linear probing approach that utilizes a pretrained Faster R-CNN to learn a highly discriminative feature representation in the penultimate layer, which is then used to train a linear classifier for novel classes. Secondly, we propose to replace the standard softmax classifier with a sigmoid classifier that is able to predict scores for each class independently, which unifies the classifier heads for both base and novel classes. Our approach outperforms strong baselines of OVOD on the COCO dataset with an $\text{AP}_{novel}$ of 40.5, setting a new state of the art. 

{\small
\bibliographystyle{ieee_fullname}
\bibliography{egbib}

\begin{thebibliography}{10}\itemsep=-1pt

\bibitem{bansal2018zero}
Ankan Bansal, Karan Sikka, Gaurav Sharma, Rama Chellappa, and Ajay Divakaran.
\newblock Zero-shot object detection.
\newblock In {\em Proceedings of the European Conference on Computer Vision (ECCV)}, pages 384--400, 2018.

\bibitem{Bravo2022locov}
M. Bravo, S. Mittal, and T. Brox.
\newblock Localized vision-language matching for open-vocabulary object detection.
\newblock In {\em German Conference on Pattern Recognition (GCPR) 2022}, 2022.

\bibitem{carion2020end}
Nicolas Carion, Francisco Massa, Gabriel Synnaeve, Nicolas Usunier, Alexander Kirillov, and Sergey Zagoruyko.
\newblock End-to-end object detection with transformers.
\newblock In {\em European conference on computer vision}, pages 213--229. Springer, 2020.

\bibitem{mmdetection}
Kai Chen, Jiaqi Wang, Jiangmiao Pang, Yuhang Cao, Yu Xiong, Xiaoxiao Li, Shuyang Sun, Wansen Feng, Ziwei Liu, Jiarui Xu, Zheng Zhang, Dazhi Cheng, Chenchen Zhu, Tianheng Cheng, Qijie Zhao, Buyu Li, Xin Lu, Rui Zhu, Yue Wu, Jifeng Dai, Jingdong Wang, Jianping Shi, Wanli Ouyang, Chen~Change Loy, and Dahua Lin.
\newblock {MMDetection}: Open mmlab detection toolbox and benchmark.
\newblock {\em arXiv preprint arXiv:1906.07155}, 2019.

\bibitem{du2022learning}
Yu Du, Fangyun Wei, Zihe Zhang, Miaojing Shi, Yue Gao, and Guoqi Li.
\newblock Learning to prompt for open-vocabulary object detection with vision-language model.
\newblock In {\em Proceedings of the IEEE/CVF Conference on Computer Vision and Pattern Recognition}, pages 14084--14093, 2022.

\bibitem{everingham2010pascal}
Mark Everingham, Luc Van~Gool, Christopher~KI Williams, John Winn, and Andrew Zisserman.
\newblock The pascal visual object classes (voc) challenge.
\newblock {\em International journal of computer vision}, 88(2):303--338, 2010.

\bibitem{fan2020few}
Qi Fan, Wei Zhuo, Chi-Keung Tang, and Yu-Wing Tai.
\newblock Few-shot object detection with attention-rpn and multi-relation detector.
\newblock In {\em Proceedings of the IEEE/CVF Conference on Computer Vision and Pattern Recognition}, pages 4013--4022, 2020.

\bibitem{feng2022promptdet}
Chengjian Feng, Yujie Zhong, Zequn Jie, Xiangxiang Chu, Haibing Ren, Xiaolin Wei, Weidi Xie, and Lin Ma.
\newblock Promptdet: Towards open-vocabulary detection using uncurated images.
\newblock In {\em Proceedings of the European Conference on Computer Vision}, 2022.

\bibitem{gao2021towards}
Mingfei Gao, Chen Xing, Juan~Carlos Niebles, Junnan Li, Ran Xu, Wenhao Liu, and Caiming Xiong.
\newblock Open vocabulary object detection with pseudo bounding-box labels.
\newblock {\em arXiv preprint arXiv:2111.09452}, 2021.

\bibitem{gu2021open}
Xiuye Gu, Tsung-Yi Lin, Weicheng Kuo, and Yin Cui.
\newblock Open-vocabulary object detection via vision and language knowledge distillation.
\newblock In {\em International Conference on Learning Representations}, 2021.

\bibitem{gupta2019lvis}
Agrim Gupta, Piotr Dollar, and Ross Girshick.
\newblock Lvis: A dataset for large vocabulary instance segmentation.
\newblock In {\em Proceedings of the IEEE/CVF conference on computer vision and pattern recognition}, pages 5356--5364, 2019.

\bibitem{he2017mask}
Kaiming He, Georgia Gkioxari, Piotr Doll{\'a}r, and Ross Girshick.
\newblock Mask r-cnn.
\newblock In {\em Proceedings of the IEEE international conference on computer vision}, pages 2961--2969, 2017.

\bibitem{he2016deep}
Kaiming He, Xiangyu Zhang, Shaoqing Ren, and Jian Sun.
\newblock Deep residual learning for image recognition.
\newblock In {\em Proceedings of the IEEE conference on computer vision and pattern recognition}, pages 770--778, 2016.

\bibitem{Huynh:CVPR22}
D. Huynh, J. Kuen, Z. Lin, J. Gu, and E. Elhamifar.
\newblock Open-vocabulary instance segmentation via robust cross-modal pseudo-labeling.
\newblock {\em {IEEE} Conference on Computer Vision and Pattern Recognition}, 2022.

\bibitem{jia2021scaling}
Chao Jia, Yinfei Yang, Ye Xia, Yi-Ting Chen, Zarana Parekh, Hieu Pham, Quoc Le, Yun-Hsuan Sung, Zhen Li, and Tom Duerig.
\newblock Scaling up visual and vision-language representation learning with noisy text supervision.
\newblock In {\em International Conference on Machine Learning}, pages 4904--4916. PMLR, 2021.

\bibitem{johnson2019billion}
Jeff Johnson, Matthijs Douze, and Herv{\'e} J{\'e}gou.
\newblock Billion-scale similarity search with {GPUs}.
\newblock {\em IEEE Transactions on Big Data}, 7(3):535--547, 2019.

\bibitem{kim2022learning}
Dahun Kim, Tsung-Yi Lin, Anelia Angelova, In~So Kweon, and Weicheng Kuo.
\newblock Learning open-world object proposals without learning to classify.
\newblock {\em IEEE Robotics and Automation Letters}, 7(2):5453--5460, 2022.

\bibitem{kuo2022f}
Weicheng Kuo, Yin Cui, Xiuye Gu, AJ Piergiovanni, and Anelia Angelova.
\newblock F-vlm: Open-vocabulary object detection upon frozen vision and language models.
\newblock {\em arXiv preprint arXiv:2209.15639}, 2022.

\bibitem{li2022dn}
Feng Li, Hao Zhang, Shilong Liu, Jian Guo, Lionel~M Ni, and Lei Zhang.
\newblock Dn-detr: Accelerate detr training by introducing query denoising.
\newblock In {\em Proceedings of the IEEE/CVF Conference on Computer Vision and Pattern Recognition}, pages 13619--13627, 2022.

\bibitem{VLDet}
Chuang Lin, Peize Sun, Yi Jiang, Ping Luo, Lizhen Qu, Gholamreza Haffari, Zehuan Yuan, and Jianfei Cai.
\newblock Learning object-language alignments for open-vocabulary object detection.
\newblock {\em arXiv preprint arXiv:2211.14843}, 2022.

\bibitem{lin2017focal}
Tsung-Yi Lin, Priya Goyal, Ross Girshick, Kaiming He, and Piotr Doll{\'a}r.
\newblock Focal loss for dense object detection.
\newblock In {\em Proceedings of the IEEE international conference on computer vision}, pages 2980--2988, 2017.

\bibitem{lin2014microsoft}
Tsung-Yi Lin, Michael Maire, Serge Belongie, James Hays, Pietro Perona, Deva Ramanan, Piotr Doll{\'a}r, and C~Lawrence Zitnick.
\newblock Microsoft coco: Common objects in context.
\newblock In {\em European conference on computer vision}, pages 740--755. Springer, 2014.

\bibitem{liu2022dabdetr}
Shilong Liu, Feng Li, Hao Zhang, Xiao Yang, Xianbiao Qi, Hang Su, Jun Zhu, and Lei Zhang.
\newblock {DAB}-{DETR}: Dynamic anchor boxes are better queries for {DETR}.
\newblock In {\em International Conference on Learning Representations}, 2022.

\bibitem{mikolov2013efficient}
Tomas Mikolov, Kai Chen, Greg Corrado, and Jeffrey Dean.
\newblock Efficient estimation of word representations in vector space.
\newblock {\em arXiv preprint arXiv:1301.3781}, 2013.

\bibitem{minderer2022simple}
Matthias Minderer, Alexey Gritsenko, Austin Stone, Maxim Neumann, Dirk Weissenborn, Alexey Dosovitskiy, Aravindh Mahendran, Anurag Arnab, Mostafa Dehghani, Zhuoran Shen, et~al.
\newblock Simple open-vocabulary object detection with vision transformers.
\newblock {\em arXiv preprint arXiv:2205.06230}, 2022.

\bibitem{pennington2014glove}
Jeffrey Pennington, Richard Socher, and Christopher~D Manning.
\newblock Glove: Global vectors for word representation.
\newblock In {\em Proceedings of the 2014 conference on empirical methods in natural language processing (EMNLP)}, pages 1532--1543, 2014.

\bibitem{qiao2021defrcn}
Limeng Qiao, Yuxuan Zhao, Zhiyuan Li, Xi Qiu, Jianan Wu, and Chi Zhang.
\newblock Defrcn: Decoupled faster r-cnn for few-shot object detection.
\newblock In {\em Proceedings of the IEEE/CVF International Conference on Computer Vision}, pages 8681--8690, 2021.

\bibitem{radford2021learning}
Alec Radford, Jong~Wook Kim, Chris Hallacy, Aditya Ramesh, Gabriel Goh, Sandhini Agarwal, Girish Sastry, Amanda Askell, Pamela Mishkin, Jack Clark, et~al.
\newblock Learning transferable visual models from natural language supervision.
\newblock In {\em International Conference on Machine Learning}, pages 8748--8763. PMLR, 2021.

\bibitem{rahman2020improved}
Shafin Rahman, Salman Khan, and Nick Barnes.
\newblock Improved visual-semantic alignment for zero-shot object detection.
\newblock In {\em AAAI}, 2020.

\bibitem{rasheed2022bridging}
Hanoona Rasheed, Muhammad Maaz, Muhammad~Uzair Khattak, Salman Khan, and Fahad~Shahbaz Khan.
\newblock Bridging the gap between object and image-level representations for open-vocabulary detection.
\newblock {\em arXiv preprint arXiv:2207.03482}, 2022.

\bibitem{redmon2017yolo9000}
Joseph Redmon and Ali Farhadi.
\newblock Yolo9000: better, faster, stronger.
\newblock In {\em Proceedings of the IEEE conference on computer vision and pattern recognition}, pages 7263--7271, 2017.

\bibitem{ren2015faster}
Shaoqing Ren, Kaiming He, Ross Girshick, and Jian Sun.
\newblock Faster r-cnn: Towards real-time object detection with region proposal networks.
\newblock {\em Advances in neural information processing systems}, 28, 2015.

\bibitem{shao2019objects365}
Shuai Shao, Zeming Li, Tianyuan Zhang, Chao Peng, Gang Yu, Xiangyu Zhang, Jing Li, and Jian Sun.
\newblock Objects365: A large-scale, high-quality dataset for object detection.
\newblock In {\em Proceedings of the IEEE/CVF international conference on computer vision}, pages 8430--8439, 2019.

\bibitem{tian2019fcos}
Zhi Tian, Chunhua Shen, Hao Chen, and Tong He.
\newblock Fcos: Fully convolutional one-stage object detection.
\newblock In {\em Proceedings of the IEEE/CVF international conference on computer vision}, pages 9627--9636, 2019.

\bibitem{pmlr-v119-wang20j}
Xin Wang, Thomas Huang, Joseph Gonzalez, Trevor Darrell, and Fisher Yu.
\newblock Frustratingly simple few-shot object detection.
\newblock In Hal~Daumé III and Aarti Singh, editors, {\em Proceedings of the 37th International Conference on Machine Learning}, volume 119 of {\em Proceedings of Machine Learning Research}, pages 9919--9928. PMLR, 13--18 Jul 2020.

\bibitem{wang2022anchor}
Yingming Wang, Xiangyu Zhang, Tong Yang, and Jian Sun.
\newblock Anchor detr: Query design for transformer-based detector.
\newblock In {\em Proceedings of the AAAI Conference on Artificial Intelligence}, volume~36, pages 2567--2575, 2022.

\bibitem{wei2021aligning}
Fangyun Wei, Yue Gao, Zhirong Wu, Han Hu, and Stephen Lin.
\newblock Aligning pretraining for detection via object-level contrastive learning.
\newblock {\em Advances in Neural Information Processing Systems}, 34:22682--22694, 2021.

\bibitem{xiao2020few}
Yang Xiao and Renaud Marlet.
\newblock Few-shot object detection and viewpoint estimation for objects in the wild.
\newblock In {\em European conference on computer vision}, pages 192--210. Springer, 2020.

\bibitem{zang2022open}
Yuhang Zang, Wei Li, Kaiyang Zhou, Chen Huang, and Chen~Change Loy.
\newblock Open-vocabulary detr with conditional matching.
\newblock {\em arXiv preprint arXiv:2203.11876}, 2022.

\bibitem{zareian2021open}
Alireza Zareian, Kevin~Dela Rosa, Derek~Hao Hu, and Shih-Fu Chang.
\newblock Open-vocabulary object detection using captions.
\newblock In {\em Proceedings of the IEEE/CVF Conference on Computer Vision and Pattern Recognition}, pages 14393--14402, 2021.

\bibitem{zhang2022dino}
Hao Zhang, Feng Li, Shilong Liu, Lei Zhang, Hang Su, Jun Zhu, Lionel~M Ni, and Heung-Yeung Shum.
\newblock Dino: Detr with improved denoising anchor boxes for end-to-end object detection.
\newblock {\em arXiv preprint arXiv:2203.03605}, 2022.

\bibitem{zhao2022exploiting}
Shiyu Zhao, Zhixing Zhang, Samuel Schulter, Long Zhao, Anastasis Stathopoulos, Manmohan Chandraker, Dimitris Metaxas, et~al.
\newblock Exploiting unlabeled data with vision and language models for object detection.
\newblock {\em arXiv preprint arXiv:2207.08954}, 2022.

\bibitem{zhong2022regionclip}
Yiwu Zhong, Jianwei Yang, Pengchuan Zhang, Chunyuan Li, Noel Codella, Liunian~Harold Li, Luowei Zhou, Xiyang Dai, Lu Yuan, Yin Li, et~al.
\newblock Regionclip: Region-based language-image pretraining.
\newblock In {\em Proceedings of the IEEE/CVF Conference on Computer Vision and Pattern Recognition}, pages 16793--16803, 2022.

\bibitem{zhou2022detecting}
Xingyi Zhou, Rohit Girdhar, Armand Joulin, Philipp Kr{\"a}henb{\"u}hl, and Ishan Misra.
\newblock Detecting twenty-thousand classes using image-level supervision.
\newblock In {\em ECCV}, 2022.

\bibitem{zhou2019objects}
Xingyi Zhou, Dequan Wang, and Philipp Kr{\"a}henb{\"u}hl.
\newblock Objects as points.
\newblock {\em arXiv preprint arXiv:1904.07850}, 2019.

\bibitem{zhu2020don}
Pengkai Zhu, Hanxiao Wang, and Venkatesh Saligrama.
\newblock Don't even look once: Synthesizing features for zero-shot detection.
\newblock In {\em Proceedings of the IEEE/CVF Conference on Computer Vision and Pattern Recognition}, pages 11693--11702, 2020.

\bibitem{zhu2020deformable}
Xizhou Zhu, Weijie Su, Lewei Lu, Bin Li, Xiaogang Wang, and Jifeng Dai.
\newblock Deformable detr: Deformable transformers for end-to-end object detection.
\newblock {\em arXiv preprint arXiv:2010.04159}, 2020.

\end{thebibliography}
}

\end{document}